\documentclass[letterpaper]{article} % DO NOT CHANGE THIS
\usepackage{aaai25}  % DO NOT CHANGE THIS
\usepackage{times}  % DO NOT CHANGE THIS
\usepackage{helvet}  % DO NOT CHANGE THIS
\usepackage{courier}  % DO NOT CHANGE THIS
\usepackage[hyphens]{url}  % DO NOT CHANGE THIS
\usepackage{graphicx} % DO NOT CHANGE THIS
\usepackage{natbib}  % DO NOT CHANGE THIS AND DO NOT ADD ANY OPTIONS TO IT
\usepackage{caption} % DO NOT CHANGE THIS AND DO NOT ADD ANY OPTIONS TO IT
\usepackage{algorithm}
\usepackage{algorithmic}

\usepackage{booktabs}
\usepackage{amsmath,amsthm,amssymb,amsfonts}
\usepackage{subfig}
\usepackage[switch]{lineno}
\usepackage{color}

\usepackage{newfloat}
\usepackage{listings}
\DeclareCaptionStyle{ruled}{labelfont=normalfont,labelsep=colon,strut=off} % DO NOT CHANGE THIS
\floatstyle{ruled}
\newfloat{listing}{tb}{lst}{}
\floatname{listing}{Listing}
\title{Reconstruction Target Matters in Masked Image Modeling for Cross-Domain Few-Shot Learning}
\author{
    Ran Ma\textsuperscript{\rm 1}, \ Yixiong Zou\textsuperscript{\rm 1}\thanks{Corresponding author.}, \ Yuhua Li\textsuperscript{\rm 1}, \ Ruixuan Li\textsuperscript{\rm 1}
}
\affiliations{
    \textsuperscript{\rm 1}School of Computer Science and Technology, Huazhong University of Science and Technology\\
    \{ranma,yixiongz,idcliyuhua,rxli\}@hust.edu.cn
}

\usepackage{bibentry}
\begin{document}

\maketitle

\begin{abstract}
Cross-Domain Few-Shot Learning (CDFSL) requires the model to transfer knowledge from the data-abundant source domain to data-scarce target domains for fast adaptation, where the large domain gap makes CDFSL a challenging problem.
Masked Autoencoder (MAE) excels in effectively using unlabeled data and learning image's global structures, enhancing model generalization and robustness. However, in the CDFSL task with significant domain shifts, we find MAE even shows lower performance than the baseline supervised models. 
In this paper, we first delve into this phenomenon for an interpretation.
We find that MAE tends to focus on low-level domain information during reconstructing pixels while changing the reconstruction target to token features could mitigate this problem. However, not all features are beneficial, as we then find reconstructing high-level features can hardly improve the model's transferability, indicating a trade-off between filtering domain information and preserving the image's global structure. 
In all, the reconstruction target matters for the CDFSL task.
Based on the above findings and interpretations, we further propose \textbf{D}omain-\textbf{A}gnostic \textbf{M}asked \textbf{I}mage \textbf{M}odeling (DAMIM) for the CDFSL task. DAMIM includes an Aggregated Feature Reconstruction module to automatically aggregate features for reconstruction, with balanced learning of domain-agnostic information and images' global structure, and a Lightweight Decoder module to further benefit the encoder's generalizability.
Experiments on four CDFSL datasets demonstrate that our method achieves state-of-the-art performance.
\end{abstract}


\section{Introduction}
With the advancement of deep learning, neural networks have become increasingly capable of handling large datasets. However, collecting large amounts of data is very difficult in some cases, such as healthcare and law enforcement, leading to the emergence of Cross-Domain Few-Shot Learning (CDFSL). This task involves first training a model on a large-scale pretraining dataset (source domain, such as a general dataset like ImageNet~\cite{imagenet}) and then transferring it to more specialized downstream datasets (target domain, such as medical datasets~\cite{ISIC}) where only a few training samples are available. Significant discrepancies between the source and target datasets usually make the transfer and downstream learning difficult ~\cite{DBLP:journals/corr/abs-1904-04232,BSCD}. Therefore, addressing domain gaps is crucial for the CDFSL task.

\begin{figure}[t]
    \centering
    \includegraphics[width=0.95\columnwidth]{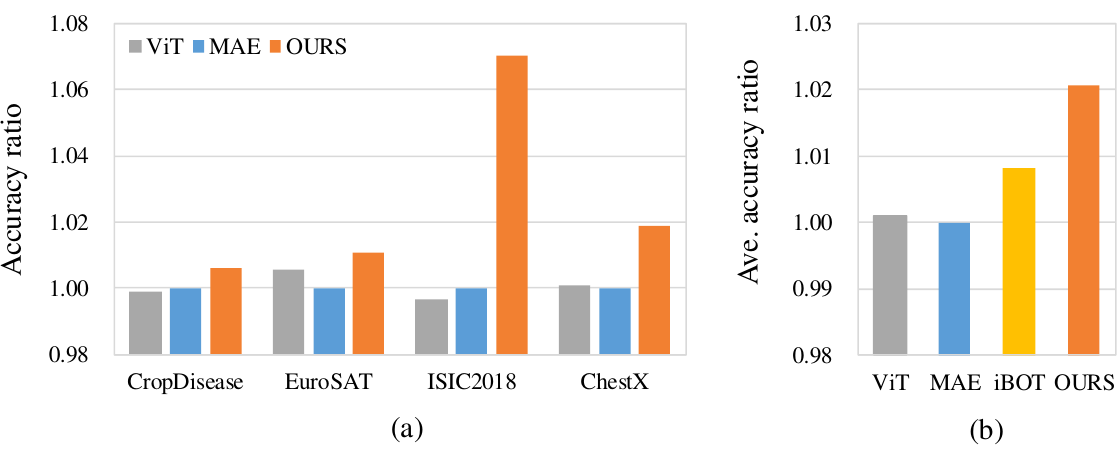}
    \vspace{-0.3cm}
    \caption{(a) The ratio of the accuracy of the supervised ViT, MAE, and our method on four CDFSL datasets, where we can see MAE underperforms on these datasets. (b) The average performance of the supervised ViT, MAE, iBOT, and our method, inspires us to think about the role of reconstruction target in MIM for CDFSL.}
    \vspace{-0.6cm}
    \label{fig:mae_poor_performance}
\end{figure}

Masked Autoencoder (MAE)~\cite{He_2022_CVPR} is a self-supervised learning method that focuses on predicting the masked parts of the image, thereby deepening the model's understanding of the image's global information. This approach has shown strong generalization capabilities across a wide range of downstream tasks~\cite{He_2022_CVPR,Kong_2023_CVPR,10.1145/3664647.3680647}. 
However, when applying MAE to train models for CDFSL tasks, we find that its performance\footnote{As the accuracy of different datasets varies largely (e.g., 90 vs. 20), we use the ratio of accuracies against MAE for illustration.} in the target domain is often unsatisfactory (Fig.~\ref{fig:mae_poor_performance}a), even lagging behind that of the Vision Transformers (ViT)~\cite{dosovitskiy2021image} trained with supervised learning. 
This counter-intuitive phenomenon makes us wonder \textit{what is it that makes MAE less effective under large domain gaps}. 

In this paper, we first delve into this phenomenon for an interpretation. We first consider Masked Image Modeling~\cite{bao2022beit} (MIM), the superset of MAE, to see whether other methods also show a similar phenomenon. We find iBOT~\cite{zhou2022image} achieves slightly higher performance than MAE under large domain gaps (Fig.~\ref{fig:mae_poor_performance}b). Since it takes token features as the target of reconstruction while MAE takes image pixels, this raises questions about the role of pixel-level reconstruction. 
Subsequently, we find MAE tends to absorb domain information during the reconstruction of image pixels, where the domain information is represented as low-level information in pixels. In contrast, shifting the reconstruction target to token features mitigates this problem, which implies the domain information gets filtered out in token features.
However, not all features are beneficial. We then find reconstructing high-level features cannot benefit transferability, indicating a trade-off between filtering out low-level domain information and preserving the image's global information.
In all, the reconstruction target matters in MIM for the CDFSL task.

Based on the above findings and interpretations,
we further propose a novel approach called \textbf{D}omain-\textbf{A}gnostic \textbf{M}asked \textbf{I}mage \textbf{M}odeling (DAMIM) for the CDFSL task. 
DAMIM incorporates two key components: the Aggregated Feature Reconstruction (AFR) module and a Lightweight Decoder (LD) module. 
The AFR module is designed to generate effective domain-agnostic reconstruction targets, thereby reducing the learning of domain-specific information that impedes generalization across different domains, while preserving the image's global information in the reconstruction.
Complementing the AFR module, since the reconstruction target is simplified from pixels to features, the LD module is introduced to prevent the encoder from relying too much on the decoder for reconstruction, thereby further improving the generalization of the feature encoder.
Comprehensive experimental analysis validates the advantages of DAMIM, with results demonstrating that it significantly improves the generalization of MIM and achieves outstanding performance on CDFSL datasets.

To summarize, our contributions are as follows:
\begin{itemize}
    \item To the best of our knowledge, we are the first to reveal the limited performance of pixel-based MIM in CDFSL.
    
    \item Through experimental analysis, we identify that the reason for the limited performance is the tendency to learn low-level domain information during reconstructing the pixels and point out that the reconstruction target matters for MIM in the CDFSL task.
    
    \item We propose Domain-Agnostic Masked Image Modeling, a novel approach that includes an Aggregated Feature Reconstruction module to automatically aggregate features for reconstruction, with balanced learning of domain-agnostic information and image's global structure, and a Lightweight Decoder module to further benefit the encoder's generalizability.
    
    \item Extensive experiments validate our analysis and methodology, demonstrating that our approach achieves state-of-the-art performance on CDFSL datasets.
\end{itemize} 

\section{Related Work}
\subsection{Cross-Domain Few-Shot Learning}
\label{sec:cdfsl_pre}
CDFSL aims to transfer knowledge from a well-trained source domain to a different target domain with limited labeled data. CDFSL is mainly studied through two approaches: transferring-based~\cite{BSCD,phoo2021STARTUP,hu2022adversarial} methods, which adapt pre-trained models from large-scale source datasets to target domains with limited data, and meta-learning~\cite{tseng2020cross,wang2021crossdomain}, which focuses on training models to quickly adapt to new tasks. In contrast to these methods, our approach emphasizes source-domain training while simultaneously enhancing knowledge transfer and target-domain fine-tuning.

\subsection{Masked Image Modeling}
MIM is a self-supervised learning method that trains models to reconstruct masked parts of an image, promoting learning valuable image representations. MIM research can be categorized by reconstruction target: pixel-based~\cite{He_2022_CVPR,Liu_2023_ICCV,Xie_2022_CVPR} and token-based~\cite{zhou2022image,SdAE,bao2022beit} methods. Pixel-based MIM methods, such as the MAE~\cite{He_2022_CVPR}, focus on reconstructing the missing pixels from the surrounding visible areas. Token-based MIM methods, in contrast, predict higher-level representations or tokens derived from the image. These methods leverage the semantic information encoded in tokens, which has shown to be highly effective in various visual recognition tasks. Despite these advances, it is important to note that no studies have specifically investigated the performance of MAE in CDFSL tasks, leaving an open area for future research.

\section{Delve into MAE on CDFSL}
In this section, we explore the reasons why MAE underperforms on CDFSL tasks with large domain gaps.

\subsection{Preliminaries}
\noindent\textbf{Cross-Domain Few-Shot Learning (CDFSL)} aims to adapt a model trained on a source domain $\mathcal{D}^S=\{(x^S_i, y^S_i)\}_{i=1}^{N_S}$ with large, diverse samples to a target domain $\mathcal{D}^T = \{(x^T_i, y^T_i)\}_{i=1}^{N_T}$, which has only a few labeled samples $N_T$ and a significant domain gap from the source domain. During adaption and evaluation on $\mathcal{D}^T$, existing research~\cite{snell2017prototypical,BSCD} employ a k-way n-shot paradigm to sample from $\mathcal{D}^T$, forming small datasets (episodes) containing $k$ classes with $n$ training samples each. The model is trained on these $k \cdot n$ samples (support set, $\{x_{ij}, y_i\}_{i=1,j=1}^{k,n}$) and then tested on $k$ classes (query set, $\{x^q\}$). This approach, alongside addressing the significant domain gap, ensures the model generalizes well to new, unseen samples in the target domain.

\begin{figure}[t]
    \centering
    \includegraphics[width=0.95\columnwidth]{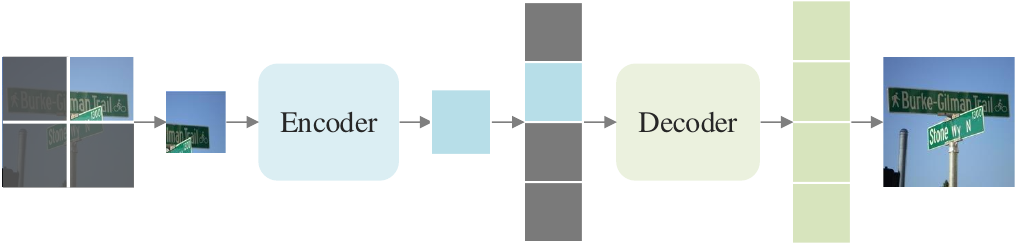}
    \caption{MAE reconstructs masked patches using an autoencoder, targeting raw pixels for reconstruction.}
    \vspace{-0.5cm}
    \label{fig:mae_structure}
\end{figure}

\vspace{0.1cm}
\noindent\textbf{Masked Autoencoder (MAE)}, as in Fig.~\ref{fig:mae_structure}, uses the architecture of an autoencoder, training the model by masking parts of the image and requiring the model to reconstruct the masked regions. Given an input image $\mathbf{X} \in \mathbb{R}^{H \times W \times C}$, following ViT ~\cite{dosovitskiy2021image}, MAE first divides the image into regular, non-overlapping patches:
\begin{equation}
    \mathbf{X} = {\{\mathbf{X}_i\}}_{i=1}^N, \quad {\mathbf{X}}_i \in \mathbb{R}^{P \times P \times C}
\end{equation}
where $N$ is the number of patches and $P$ is the patch size. Then, with a mask ratio $r$, some patches are randomly masked. The mask is defined as a vector $\mathbf{m} \in \{0,1\}^N$, where $N \times r$ elements are randomly set to 0, and the remaining elements are set to 1, we use $\mathbf{m}$ to mask patches:
\begin{equation}
    \mathbf{X}^{vis} = \mathbf{X} \odot \mathbf{m}.
\end{equation}
$\mathbf{X}^{vis} = [\mathbf{x_1}, \mathbf{x_2}, \dots, \mathbf{x_{N\times (1-r)}}]$ is the visible patches set, and $\mathbf{X}^{mask} = [\mathbf{x_{N\times (1-r)+1}}, \mathbf{x_{N\times (1-r)+2}}, \dots, \mathbf{x_N}]$ denotes the masked patches. The encoder $f_{enc}$ processes visible patches $\mathbf{X}^{vis}$ as input and outputs the representations:
\begin{equation}
    Z = f_{enc}(\mathbf{X}^{vis}), \quad Z \in \mathbb{R}^{(N \times (1-r)) \times d}
\end{equation}
where $d$ is the representation dimension. Then the decoder $f_{dec}$ receives $Z$ and the encoded mask patches $M \in \mathbb{R}^{(N \times r) \times d}$, and outputs the reconstructed image:
\begin{equation}
    \hat{\mathbf{X}} = f_{dec}([Z,M]), \quad \hat{\mathbf{X}} = [\hat{\mathbf{x}}_1, \hat{\mathbf{x}}_2, \dots, \hat{\mathbf{x}}_N].
\end{equation}
The model is trained by minimizing the difference between the reconstructed image and the original image, typically using the Mean Squared Error (MSE) as the loss function:
\begin{equation}
    \mathcal{L} = \frac{1}{N \times r} \sum_{i \in \mathbf{X}^{mask}} \| \hat{\mathbf{x}}_i - \mathbf{x}_i \|_2^2
\end{equation}
In this paper, we use unsupervised MAE to train the encoder on the source domain without requiring labels. Then we discard the decoder, retain the encoder as the backbone, and use ProtoNet~\cite{liu2020prototype} with a distance function $d(\cdot,\cdot)$ for the target domain few-shot evaluation:
\begin{equation}
    \hat{y_q} = arg \min_{i} d(\frac{1}{n} \sum_j f_{enc}(x_{ij}), f_{enc}(x_q)).
    \label{eq:prototype_classification}
\end{equation}

\subsection{Delve into MAE's limited performance in CDFSL}
\subsubsection{MAE tends to learn low-level information.}
\begin{figure}[t]
    \centering
    \includegraphics[width=0.95\columnwidth]{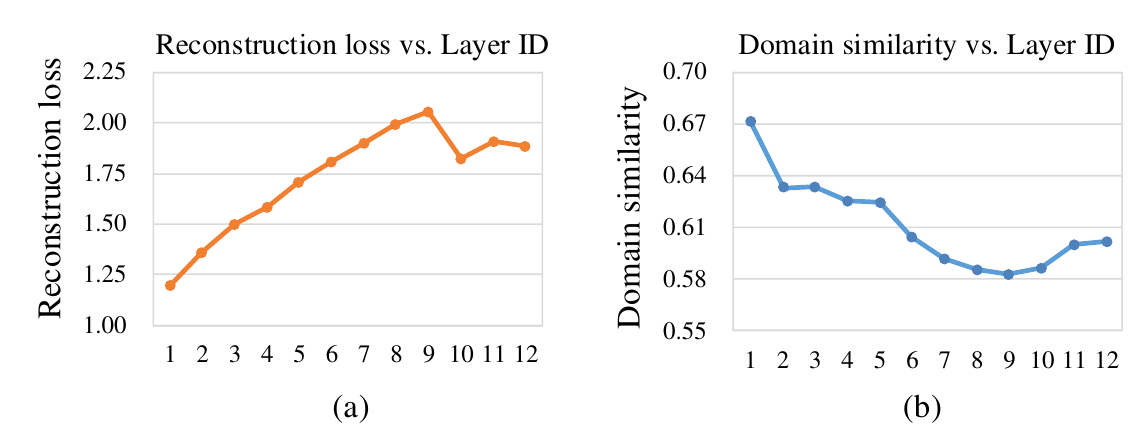}
    \vspace{-0.3cm}
    \caption{(a) Reconstruction loss measured by using different layer features in ViT as reconstruction targets. The reconstruction loss of shallow-layer features is lower, indicating it is easier for the model to capture and learn low-level features. (b) Domain similarity of the final features between the source and target domains after disrupting features in different layers. Disrupting shallow-layer features leads to a higher domain similarity.}
    \label{fig:analysis1}
\end{figure}

We observe that the token-based iBOT~\cite{zhou2022image} model outperforms the pixel-based MAE in the CDFSL task, which inspires us to think about the role of pixel-level reconstruction. Specifically, we hypothesize that MAE's focus on pixel reconstruction might contribute to its ineffectiveness in CDFSL tasks,
because pixel reconstruction requires the model to learn low-level visual information, such as the brightness or color of images that could be relevant to domains.
To investigate the learning behavior of MAE during reconstruction, we set different features in the ViT as the reconstruction target. As shown in Fig.~\ref{fig:analysis1}(a), the reconstruction loss for shallow-layer features is significantly lower than that for deeper-layer ones, indicating that the model is more inclined to capture the low-level information. This tendency might lead the MAE to prioritize learning low-level information over more semantic and global representations. 

\begin{figure}[t]
    \centering
    \includegraphics[width=0.95\columnwidth]{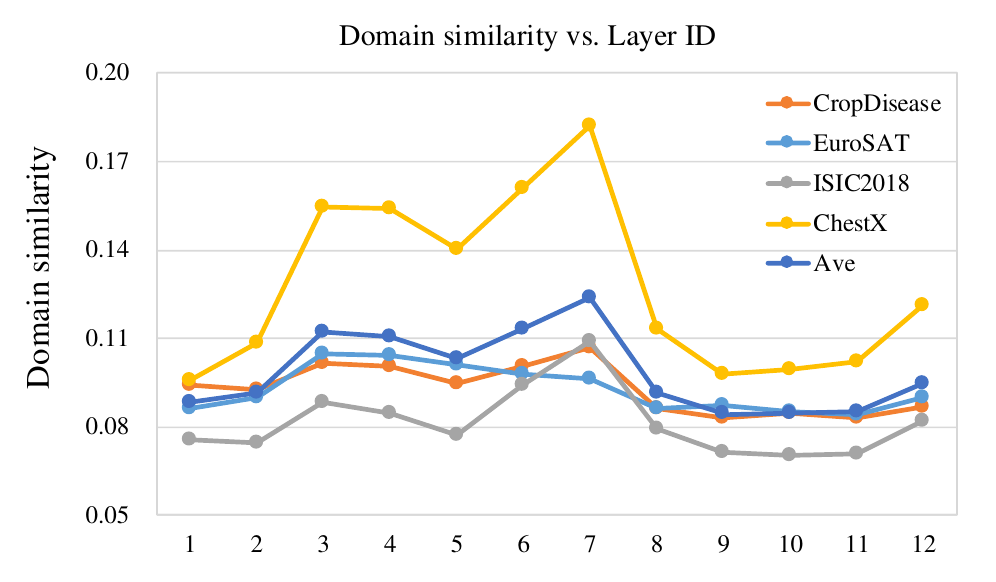}
    \vspace{-0.3cm}
    \caption{Domain similarity of the models using different layer features in ViT as reconstruction targets. Shallow-layer features show lower domain similarity, while reconstructing deeper-layer features can hardly improve the model's transferability, indicating a trade-off between filtering domain information and preserving the image's global structure.}
    \vspace{-0.3cm}
    \label{fig:analysis2}
\end{figure}

\begin{figure}[t]
    \centering
    \includegraphics[width=0.95\columnwidth]{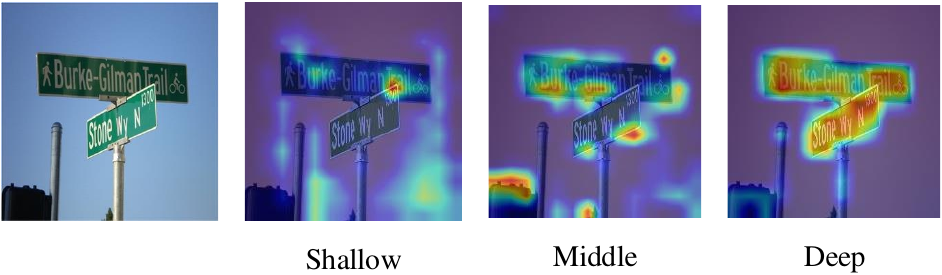}
    \vspace{-0.3cm}
    \caption{Shallow, middle, and deep features visualization of MAE. In shallow layers, the model predominantly captures low-level information. As the network goes deeper, its focus gradually shifts towards semantic parts in images.}
    \vspace{-0.5cm}
    \label{fig:analysis3}
\end{figure}

%%%%%%%%%%%%%% figure: our model %%%%%%%%%%%%
\begin{figure*}[t] 
    \centering
    \includegraphics[width=1.9\columnwidth]{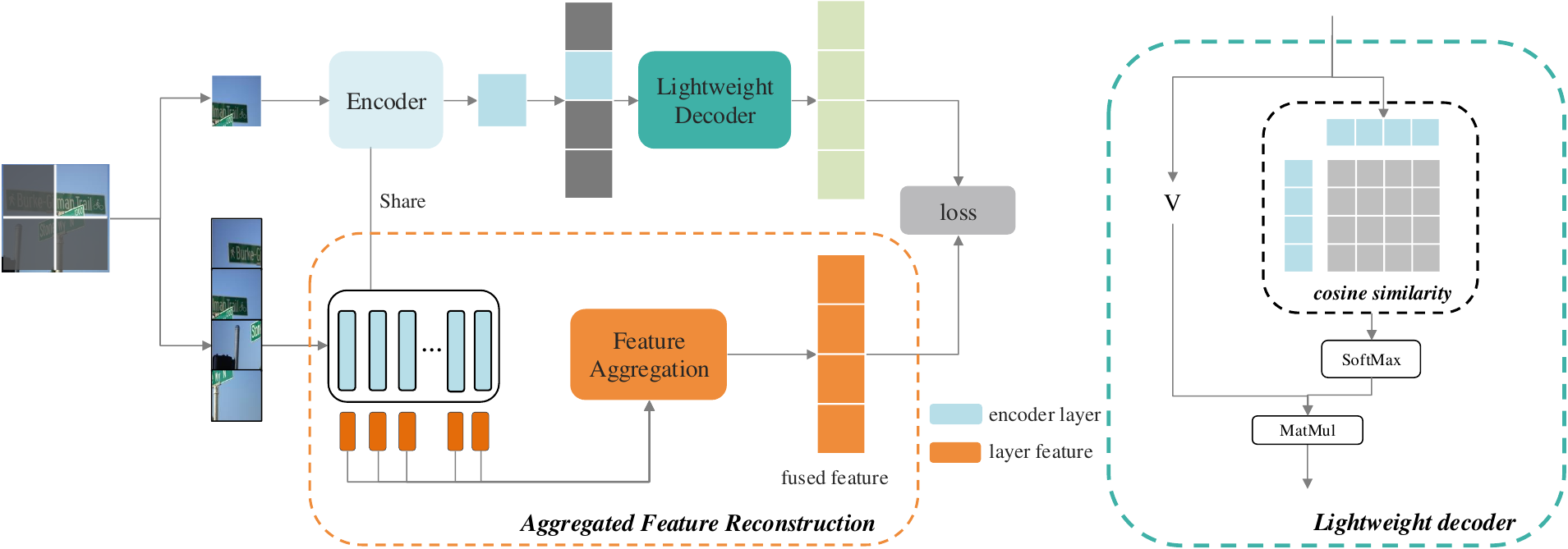}
    \caption{Our proposed \textbf{DAMIM} model comprises two key components: (a) The Aggregated Feature Reconstruction (AFR) module (in the orange box), which integrates multi-layer features as reconstruction targets. (b) The Lightweight Decoder (LN) module (in the green box) substitutes the original query-key attention in ViT with a cosine similarity in a simplified structure.}
    \vspace{-0.3cm}
    \label{fig:our_model}
\end{figure*}
%%%%%%%%%%%%%% figure: our model %%%%%%%%%%%%

\begin{figure}[t]
    \centering
    \includegraphics[width=0.95\columnwidth]{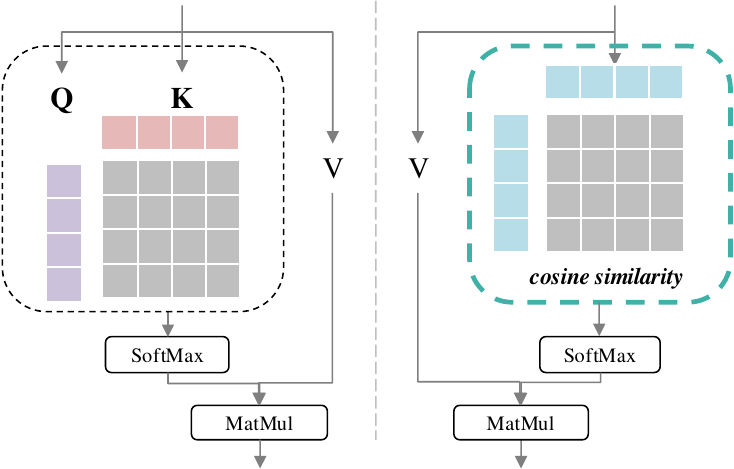}
    \vspace{-0.3cm}
    \caption{Our Lightweight Decoder. Compared to the query-key mechanism, LD uses a simple cosine similarity score between tokens, reducing computational cost while still effectively capturing essential relationships.}
    \vspace{-0.3cm}
    \label{fig:lightweight_decoder}
\end{figure}

\subsubsection{Low-level information exhibits domain specificity.} To delve into the properties of low-level information, we apply a masking operation to disrupt the output of different ViT layers. Then, we measure the domain similarity by the CKA\footnote{Please see the appendix for details.} similarity~\cite{kornblith2019similarity} of the final output features between the source and target domains, as shown in Fig.~\ref{fig:analysis1}(b). 
As can be seen, when low-level information is disrupted, the domain similarity increases significantly, compared to when higher-level features are disrupted. This finding verifies that low-level information is highly domain-specific, containing rich domain information closely tied to the source domain. 
%If the lower-level information heavily reflects the source domain patterns, disrupting this information will increase the similarity between the source and target domains. 
Therefore, we can conclude that MAE tends to prioritize lower-level information during the reconstruction process, and the domain information carried by it hampers the model's ability to generalize effectively from the source domain to the target domain.
On the other hand, with the increase in the depth of the network, such information is gradually filtered out, which makes the model capture more semantic information than low-level information. As a result, it might improve the model's generalizability to replace pixels with token features as the reconstruction target.

\subsubsection{Reconstruction with high-level features.}Since pixel-level or low-level features are rich in domain-specific information, we are motivated to explore using higher-level features as the reconstruction targets. As shown in Fig.~\ref{fig:analysis2}, when different layer features are used as reconstruction targets, the low-level features exhibit lower domain similarity between the source and target domains, consistent with the findings discussed in the previous section. 
However, not all features at different levels are beneficial to domain generalization, 
%especially when there is a significant domain gap. 
especially for the higher-level features near the output of the network.
In the shallow layers, the model predominantly captures low-level information, such as color distribution and brightness, which are often domain-specific and difficult to transfer effectively. As the network goes deeper, its focus gradually shifts towards more abstract and semantic information, as depicted in Fig.~\ref{fig:analysis3}. 
%These higher-level features are less transferable but crucial for the model's learning process. 
These higher-level features tend to activate only the regions for semantic objects, with less focus on other background regions, limiting MIM's ability to learn the image's global structure.
Therefore, higher-level features result in a sharp decline in domain similarity. Consequently, there is a trade-off between filtering out low-level domain-specific information and preserving the image's global information.

\subsection{Conclusion and Discussion}

The reconstruction target matters for MIM in the CDFSL task. For low-level features or pixels, domain information is carried as the low-level information (e.g., color distribution, brightness, etc.) that harms the generalization. For high-level features, the model tends to focus on only the semantic parts in the image, harming MIM's ability to learn the image's global structure. Therefore, methods need to be designed for the automatic selection of the reconstruction target.

\section{Method}
In this section, we introduce our Domain-Agnostic Masked Image Modeling (DAMIM) approach for CDFSL, as illustrated in Fig.~\ref{fig:our_model}. Firstly, we propose a Feature Aggregation Reconstruction (AFR) module, which integrates features from multiple layers to serve as the reconstruction target. To complement AFR, we design a Lightweight Decoder (LD) module that prevents the encoder from over-relying on the decoder for reconstruction, thereby enhancing the generalization of the feature encoder. The following sections detail each component of our model.

\subsubsection{Aggreated Feature Reconstruction Module}
To balance filtering low-level domain information while preserving global information, we propose the AFR Module, automatically aggregating the most advantageous features as a reconstruction target to achieve this balance. 

Firstly, we establish an auxiliary encoder that shares weights with the original encoder to extract features from multiple layers. The original encoder processes only the visible patches of an image, while the auxiliary encoder processes all patches. Following MAE~\cite{He_2022_CVPR}, the input image $\mathbf{X} \in \mathbb{R}^{H \times W \times C}$ is divided into non-overlapping patches $\{\mathbf{X}_1, \mathbf{X}_2, \dots, \mathbf{X}_N\}$, where each patch $\mathbf{X}_i \in \mathbb{R}^{P \times P \times C}$. A high proportion of these patches is then randomly masked, resulting in a visible patch set $\mathbf{X}^{vis} = [\mathbf{x}_1, \mathbf{x}_2, \dots, \mathbf{x}_{N^{\prime}}]$ and a masked patch set $\mathbf{X}^{mask} =[\mathbf{x}_{N^{\prime}+1}, \mathbf{x}_{N^{\prime}+2}, \dots, \mathbf{x}_N]$, where $N^{\prime}$ is the number of visible patches. The original encoder processes the visible patches to obtain their latent representations:
\begin{equation}
    \mathcal{Z} = \mathbf{Encoder}(\mathbf{X}^{vis}), \quad \mathcal{Z} \in \mathbb{R}^{N^{\prime} \times d}.
\end{equation}
Simultaneously, the auxiliary encoder processes full image patches and outputs features at the $l$-th layer:
\begin{equation}
    f^{(l)} = \mathbf{Encoder}_{aux}^{(l)}(\mathbf{X}), \quad f^{(l)} \in \mathbb{R}^{N \times d}.
\end{equation}
$f^{(l)}$ represents the output of the $l$-th layer of the auxiliary encoder $\mathbf{Encoder}_{aux}^{(l)}$.
Then, we introduce a feature aggregation module to fuse layer-wise features, and a projection layer aligns the feature spaces across different levels. For a given layer $l$, the feature alignment is achieved by:
\begin{equation}
    \tilde{f}^{(l)} = W^{(l)} f^{(l)}, \quad W^{(l)} \in \mathbb{R}^{d \times d}
\end{equation}
where $W^{(l)}$ is the projection matrix for the $l$-th layer. To combine these aligned features, we use a weighted average pooling mechanism defined as:
\begin{equation}
    \mathcal{F} = \sum^{L}_{l=1}\alpha^{(l)}\tilde{f}^{(l)}, \mathcal{F} \in \mathbb{R}^{N \times d}.
    \label{feature_reconstruction}
\end{equation}
where $\alpha^{(l)}$ are feature weights, which are generated based on the layer reconstruction loss with a linear and softmax layer. The aggregated feature ${\mathcal{F}}$ serves as the reconstruction target, enabling effective feature combination across layers.

\subsubsection{Lightweight Decoder}
Since the reconstruction target is feature-based, we design a lightweight decoder to prevent the encoder from over-relying on the decoder during the MAE reconstruction process, ensuring the encoder focuses on learning more generalizable features. 

Our decoder consists of a single Transformer block with just one attention head, significantly reducing computational complexity. Instead of the conventional query-key attention mechanism~\cite{NIPS2017_3f5ee243}, we use a simple cosine similarity score between tokens, as shown in Fig.~\ref{fig:lightweight_decoder}:
\begin{equation}
    \text{CosSim}_{ij} = \frac{\mathbf{t}_i \cdot \mathbf{t}_j}{\|\mathbf{t}_i\| \|\mathbf{t}_j\|}, \quad i,j = 1, \dots, N
\end{equation}
where $\mathbf{t}_i$ denotes an image token. This substitution not only simplifies the architecture but also better aligns with our goal of reducing the decoder’s influence. We also remove the MLP (Multi-Layer Perceptron) module entirely, further streamlining the decoder. We then concatenate the visible representations $\mathcal{Z}$, processed by the encoder, with the mask tokens $M \in \mathbb{R}^{(N - N^{\prime}) \times d}$, which are obtained by embedding the mask patches and adding positional information:
\begin{equation}
    \begin{aligned}
        \mathcal{R} = \mathbf{Decoder}({\mathcal{Z}}, M), \quad \mathcal{R} \in \mathbb{R}^{n \times d}
    \end{aligned}
\end{equation}
The processed output from our decoder yields the final reconstruction $\mathcal{R}$.

\subsubsection{Reconstruction Loss}
The final reconstruction loss $\mathcal{L}_{recon}$ is calculated using the Mean Squared Error (MSE) between the reconstructed output $\mathcal{R}$ and the target aggregated features $\mathcal{F}$, which is defined as:
\begin{equation}
    \mathcal{L}_{recon} = \frac{1}{N} \sum_{i=1}^{N} 
    \left\| {\mathcal{F}}_i - {\mathcal{R}}_i \right\|^2
\end{equation}

\subsubsection{Target domain evaluation and fine-tuning}During the evaluation and fine-tuning phase in the target domain, the decoder is discarded, and the encoder is retained as the backbone. Following recent works~\cite{Shell2023Pushing,styleadv}, we perform prototype-based classification within few-shot episodes as described in Eq.\eqref{eq:prototype_classification} or fine-tune the backbone using the support set from these episodes for the classifier-based classification.

%%%%%%%%%%%%%%%%% 1-shot sota %%%%%%%%%%%%%%%%%%
\begin{table*}[t]
  \centering
    
    \resizebox{1.9\columnwidth}{!}{
    \begin{tabular}{cccccccc}
    \toprule
    Method & Mark  & FT    & ChestX & ISIC2018 & EuroSAT & CropDisease & Average \\
    \midrule
    MEM-FS~\cite{MEM-FS} & TIP-23 & $\times$ & 22.76  & 32.97  & 68.11  & 81.11  & 51.24  \\
    StyleAdv~\cite{styleadv} & CVPR-23 & $\times$ & 22.92  & 33.05  & 72.15  & 81.22  & 52.34  \\
    SDT~\cite{LIU2024106536} & NN-24 & $\times$ & 22.79 & 33.40 & 72.71 & 81.03 & 52.48 \\
    FLoR~\cite{zou2024flatten}  & CVPR-24 & $\times$ & 22.78  & 34.20  & 72.39  & 81.81  & 52.80  \\
    \textbf{Ours} & \textbf{Ours} & $\times$ & \textbf{22.97 } & \textbf{34.66 } & \textbf{72.87 } & \textbf{82.34 } & \textbf{53.21 } \\
    \midrule
    PMF~\cite{Shell2023Pushing} & CVPR-22 & $\checkmark$ & 21.73  & 30.36  & 70.74  & 80.79  & 50.91 \\
    FLoR~\cite{zou2024flatten}  & CVPR-24 & $\checkmark$ & 23.26  & 35.49  & 73.09  & 83.55  & 53.85 \\
    StyleAdv~\cite{styleadv} & CVPR-23 & $\checkmark$ & 22.92  & 33.99  & \textbf{74.93 } & \textbf{84.11 } & 53.99  \\
    \textbf{Ours} & \textbf{Ours} & $\checkmark$ & \textbf{23.38} & \textbf{36.35} & \underline{73.61}  & \underline{83.90}  & \textbf{54.31} \\
    \midrule
    MEM-FS+RDA\textsuperscript{*}~\cite{MEM-FS} & TIP-23 & $\checkmark$ & 23.85  & 37.07  & 75.91  & 83.74  & 55.14  \\
    \textbf{Ours\textsuperscript{*}} & \textbf{Ours} & $\checkmark$ & \textbf{23.91} & \textbf{38.07} & \textbf{77.23} & \textbf{86.74} & \textbf{56.49} \\
    \bottomrule
    \end{tabular}}
    \vspace{-0.3cm}
    \caption{Comparison with the state-of-the-art works based on ViT-S by 5-way 1-shot accuracy.}
    \label{tab:1shot_sota}
\end{table*}%
%%%%%%%%%%%%%%%%% 1-shot sota %%%%%%%%%%%%%%%%%%

%%%%%%%%%%%%%%%%% 5-shot sota %%%%%%%%%%%%%%%%%%
\begin{table*}[t]
  \centering
    
    \resizebox{1.9\columnwidth}{!}{
    \begin{tabular}{cccccccc}
    \toprule
    Method & Mark & FT & ChestX & ISIC2018 & EuroSAT & CropDisease & Average \\
    \midrule
    MEM-FS~\cite{MEM-FS} & TIP-23 & $\times$ & 26.67  & 47.38  & 86.49  & 93.74  & 63.57  \\
    StyleAdv~\cite{styleadv} & CVPR-23 & $\times$ & 26.97  & 47.73  & 88.57  & 94.85  & 64.53  \\
    SDT~\cite{LIU2024106536} & NN-24 & $\times$ & 26.72 & 47.64 & 89.60 & 95.00 & 64.75 \\
    FLoR~\cite{zou2024flatten} & CVPR-24 & $\times$ & 26.71  & 49.52  & \textbf{90.41 } & 95.28  & 65.48  \\
    \textbf{Ours} & Ours  & $\times$ & \textbf{27.28 } & \textbf{50.76 } & 89.50  & \textbf{95.52 } & \textbf{65.77 } \\
    \midrule
    PMF~\cite{Shell2023Pushing} & CVPR-22 & $\checkmark$ & 27.27  & 50.12  & 85.98  & 92.96  & 64.08  \\
    StyleAdv~\cite{styleadv} & CVPR-23 & $\checkmark$ & 26.97  & 51.23  & 90.12  & 95.99  & 66.08  \\
    FLoR~\cite{zou2024flatten}  & CVPR-24 & $\checkmark$ & 27.02  & 53.06  & 90.75  & \textbf{96.47 } & 66.83  \\
    \textbf{Ours} & \textbf{Ours} & $\checkmark$ & \textbf{27.82 } & \textbf{54.86 } & \textbf{91.18 } & \underline{96.34}  & \textbf{67.55} \\
    \midrule
    MEM-FS+RDA\textsuperscript{*}~\cite{MEM-FS} & TIP-23 & $\checkmark$ & 27.98  & 51.02  & 88.77  & 95.04  & 65.70  \\
    \textbf{Ours\textsuperscript{*}} & \textbf{Ours} & $\checkmark$ & \textbf{28.10} & \textbf{55.44} & \textbf{91.08} & \textbf{96.49} & \textbf{67.78} \\
    \bottomrule
    \end{tabular}}
    \vspace{-0.3cm}
    \caption{Comparison with the state-of-the-art works based on ViT-S by 5-way 5-shot accuracy.}
    \label{tab:5shot_sota}
    \vspace{-0.3cm}
\end{table*}

%%%%%%%%%%%%%%%%% Experiments %%%%%%%%%%%%%%%%%%
\section{Experiments}

\begin{figure}[t] 
    \centering
    \includegraphics[width=0.95\columnwidth]{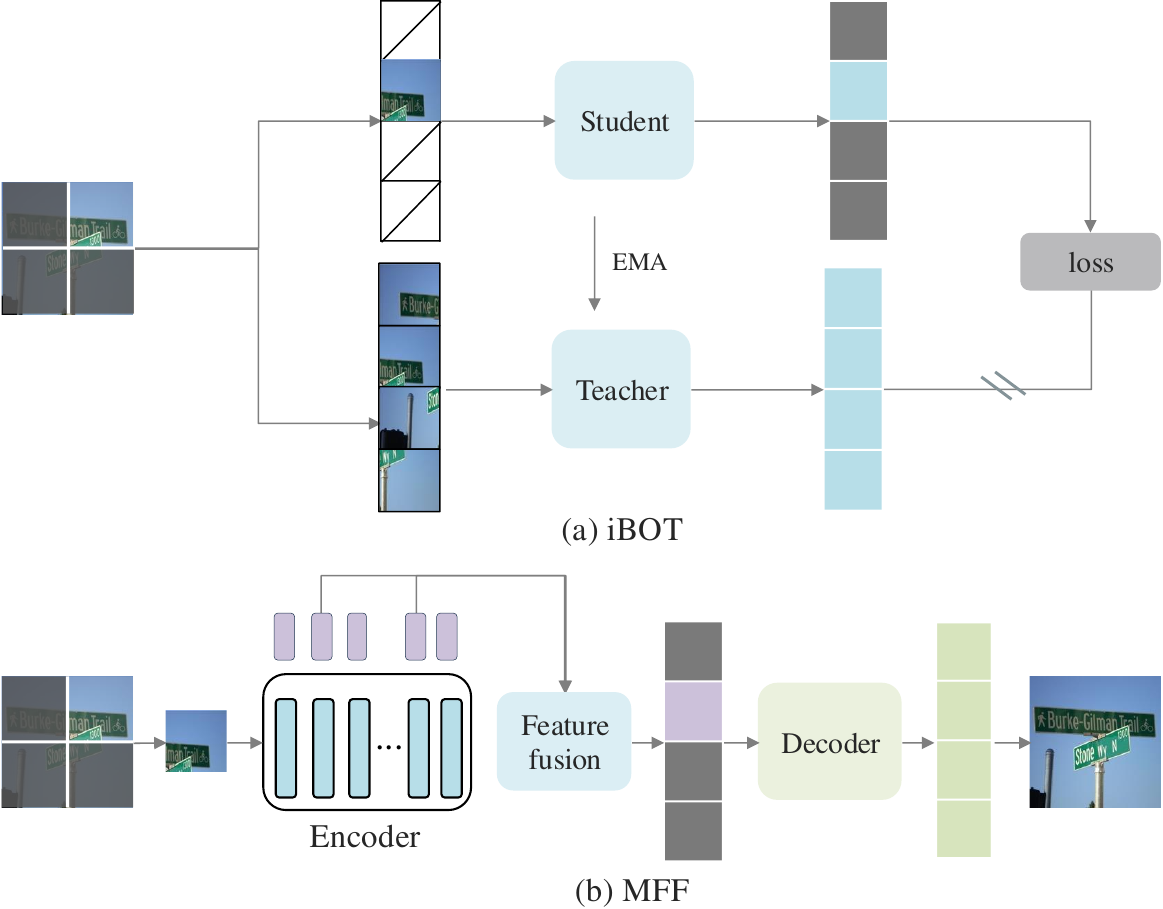}
    \caption{The structures of iBOT and MFF. iBOT uses a teacher-student framework with mask tokens visible in the encoder, while DAMIM employs a lightweight decoder for reconstruction. MFF fuses features before the decoder, while DAMIM uses aggregated features as reconstruction targets.}
    \label{fig:ibot_mff}
\end{figure}

%%%%%%%%%%%%%%%%% Dataset %%%%%%%%%%%%%%%%%%
\subsection{Dataset}
Following BSCD benchmark~\cite{BSCD}, we utilize miniImagenet~\cite{Vinyals2016Matching} as the source dataset and four cross-domain datasets as the target datasets, including CropDisease~\cite{34699834fa624a3bbc2fae48eb151339}, EuroSAT\cite{helber2019eurosat}, ISIC2018~\cite{ISIC} and ChestX~\cite{Wang_2017} for few-shot training and evaluation. MiniImageNet includes 100 categories of natural images, each with 600 images, split into 64 training categories, 16 validation categories, and 20 test categories. CropDisease includes 38 categories of crop diseases with 43,456 images. EuroSAT is used for land use and cover classification from satellite images, containing 27,000 images across 10 categories. ISIC2018 comprises 10,015 skin lesion images in 7 categories. ChestX focuses on thoracic disease diagnosis, with 25,847 images across 7 categories.

%%%%%%%%%%%%%%%%% Implementation Details %%%%%%%%%%%%%%%%%%
\subsection{Implementation Details}
We follow StyleAdv~\cite{styleadv} to take ViT-S as the backbone network and take the DINO~\cite{caron2021emerging} pretraining on ImageNet1K~\cite{imagenet} as the initialization. During the base class training, our model is trained with the AdamW optimizer~\cite{loshchilov2019decoupled} with a learning rate of 0.001 for the classifier, 1e-7 for the backbone network, and 1e-6 for our decoder. During the novel-class fine-tuning, we discard the decoder and only fine-tune the encoder backbone. We use the SGD optimizer with a momentum of 0.9 to fine-tune all parameters.

\begin{figure}[t] 
    \centering
    \includegraphics[width=0.95\columnwidth]{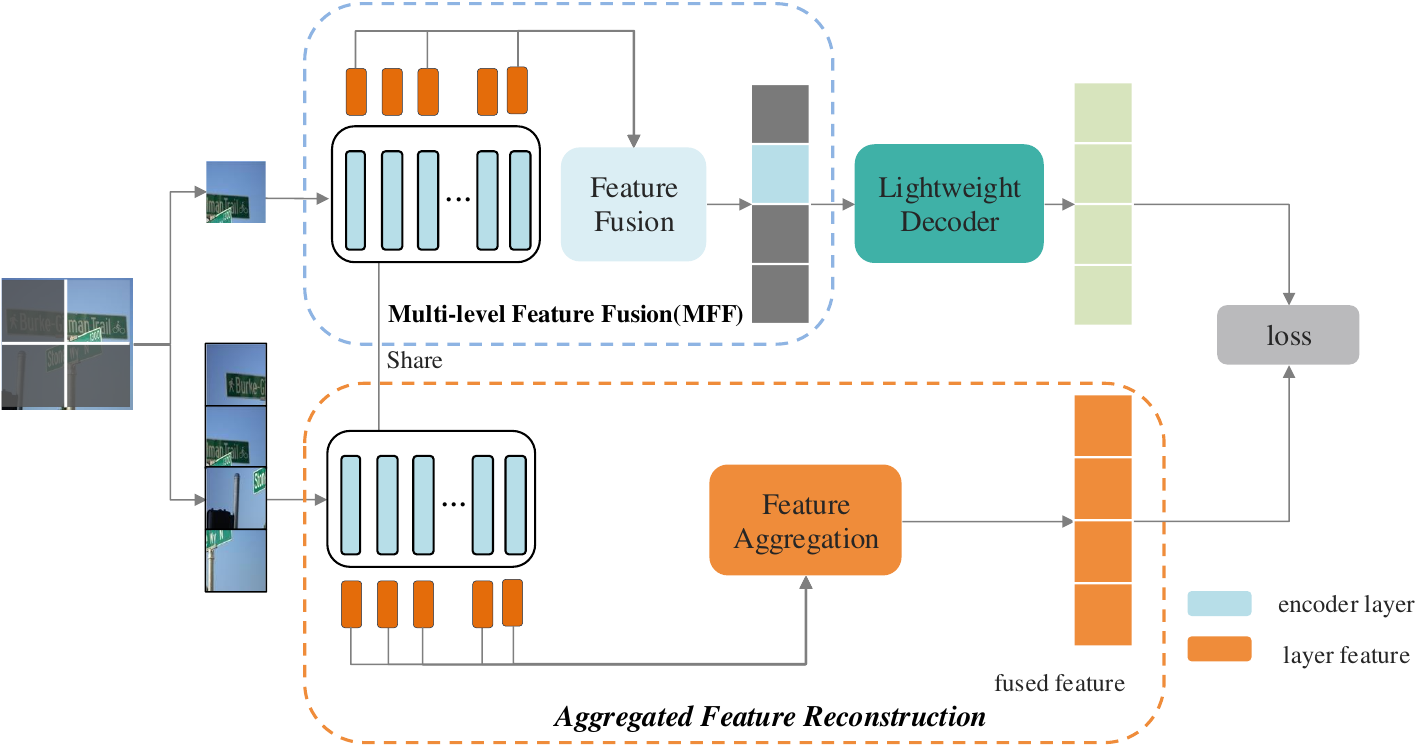}
    \caption{Applying our method to MFF. MFF improves upon the pixel-based MAE by fusing features from multiple encoder blocks before the decoder. Our approach explicitly avoids learning domain-specific features by using aggregated multi-layer features as the reconstruction target.}
    \label{fig:ourmodel+mff}
\end{figure}

%%%%%%%%%%%%%%%%% comparison with other MIM methods %%%%%%%%%%%%%%%%%%
\begin{table}[t]
    \centering
    \resizebox{0.95\columnwidth}{!}{
    \begin{tabular}{cccccc}
        \toprule
        Method & CropDisease & EuroSAT & ISIC2018 & ChestX & Ave. \\
        \midrule
        iBOT & 94.94 & \textbf{89.58} & 48.34 & 26.97 & 64.96 \\
        MFF & 95.15 & 89.16 & 49.83 & 27.22 & 65.34 \\
        \textbf{Ours} & \textbf{95.52 } & \underline{89.50} & \textbf{50.76} & \textbf{27.28} & \textbf{65.77 } \\
        \midrule
        iBOT & 94.94 & 89.58 & 48.34 & 26.97 & 64.96 \\
        \textbf{iBOT+Ours} & \textbf{95.30} & 89.57 & \textbf{48.67} & \textbf{27.02} & \textbf{65.14}\\
        \midrule
        MFF & 95.15 & 89.16 & 49.83 & 27.22 & 65.34 \\
        \textbf{MFF+Ours} & \textbf{95.28} & \textbf{89.51} & \textbf{51.21} & \textbf{27.85} & \textbf{65.96} \\
        \bottomrule
    \end{tabular}}
    \vspace{-0.3cm}
    \caption{Comparison with iBOT and MFF by 5-way 5-shot accuracy. DAMIM outperforms both methods, further improving performance and achieving a new state-of-the-art.}
    \label{tab:comparison_mim}
\end{table}

\begin{table}[t]
    \centering
    \resizebox{0.95\columnwidth}{!}{
    \begin{tabular}{ccccccc}
        \toprule
        AFR & LD & CropDisease & EuroSAT & ISIC2018 & ChestX & Ave. \\
        \midrule
        & & 94.88 & 86.94 & 46.00 & 26.48 & 63.58 \\
        $\checkmark$ & & 95.25 & 88.63 & 49.87 & 26.58 & 65.08 \\
        & $\checkmark$ & 95.21 & 88.90 & 49.74 & 27.10 & 65.24 \\
        $\checkmark$ & $\checkmark$ & \textbf{95.52} & \textbf{89.50} & \textbf{50.38} & \textbf{27.28} & \textbf{65.77} \\
        \bottomrule
    \end{tabular}}
    \vspace{-0.3cm}
    \caption{Ablation study by the 5-way 5-shot accuracy.}
    % \vspace{-0.6cm}
    \label{tab:ablation}
\end{table}

\begin{table}[t]
    \centering
    \resizebox{0.95\columnwidth}{!}{
    \begin{tabular}{cccccc}
        \toprule
        Method & CropDisease & EuroSAT & ISIC2018 & ChestX & Ave. \\
        \midrule
        BL & 0.0982 & 0.0907 & 0.0705 & 0.0858 & 0.0863 \\
        BL+AFR & 0.1071 & 0.0961 & 0.1091 & 0.1822 & 0.1236 \\
        BL+AFR+LD & 0.1305 & 0.1226 & 0.1254 & 0.2252 & 0.1509 \\
        \bottomrule
    \end{tabular}}
    \vspace{-0.3cm}
    \caption{Domain similarity of our method. Both the AFR and LD improve domain similarity in CDFSL tasks.}
    \label{tab:feaRecon_cka}
\end{table}

%%%%%%%%%%%%%%%%% Comparison with SOTA Method %%%%%%%%%%%%%%%%%%
\subsection{Comparison with SOTA Method}
Tab.~\ref{tab:1shot_sota} and Tab.~\ref{tab:5shot_sota} compare our results with state-of-the-art methods using the ViT-S backbone pretrained by DINO in 1-shot and 5-shot settings. We distinguish results with and without fine-tuning (FT), with an asterisk (*) indicating a transductive setting. Comparisons with CNN backbone methods are provided in the supplementary materials. PMF~\cite{Shell2023Pushing}, StyleAdv~\cite{styleadv}, MEM-FS~\cite{MEM-FS}, SDT~\cite{LIU2024106536} and FLoR~\cite{zou2024flatten} are introduced as competitors. Our consistently superior performance across all settings highlights the effectiveness of our DAMIM.

\subsection{Comparison with other MIM Methods}
We compare DAMIM with other MIM models, specifically the token-based iBOT~\cite{zhou2022image} and the pixel-based MFF~\cite{Liu_2023_ICCV}, as shown in Tab.~\ref{tab:comparison_mim}. The structures of these two models are depicted in Fig.~\ref{fig:ibot_mff}. iBOT uses a teacher-student framework, where the student reconstructs masked patches using the teacher's output. The mask tokens are visible to the entire encoder in iBOT, while our method uses a lightweight decoder for reconstruction. MFF enhances MAE by fusing features from multiple encoder blocks before the decoder, which implicitly prevents overfitting to low-level features. In contrast, our method explicitly avoids learning low-level domain information by using aggregated multilayer features as the reconstruction target. The results show that DAMIM outperforms both iBOT and MFF. Additionally, applying our method to iBOT and MFF further boosts performance. Notably, combining our approach with MFF, as shown in Fig.~\ref{fig:ourmodel+mff}, achieved a new state-of-the-art, highlighting the effectiveness and adaptability of our method.

%%%%%%%%%%%%%%% Ablation Study %%%%%%%%%%%%%%%%
\subsection{Ablation Study}
%%%%%%%%%%%%%%% method ablation table %%%%%%%%%%%%%%%%

% \begin{table}[t]
%     \centering
%     \resizebox{0.95\columnwidth}{!}{
%     \begin{tabular}{cccccc}
%         \toprule
%         Method & CropDisease & EuroSAT & ISIC2018 & ChestX & Ave. \\
%         \midrule
%         BL+AFR & 0.1071 & 0.0961 & 0.1091 & 0.1822 & 0.1236 \\
%         BL+AFR+LD & 0.1305 & 0.1226 & 0.1254 & 0.2252 & 0.1509 \\
%         \bottomrule
%     \end{tabular}}
%     \vspace{-0.3cm}
%     \caption{Domain similarity for baseline with AFR and combined with LD. LD further enhances domain similarity.}
%     \label{tab:decoder_cka}
%     \vspace{-0.3cm}
% \end{table}

\vspace{0.1cm}
\noindent\textbf{Vertification of DAMIM}
We conduct an ablation study on DAMIM, as shown in Tab.~\ref{tab:ablation}. The results demonstrate that both the Aggregated Feature Reconstruction module and the Lightweight Decoder module independently improve the baseline by 1.50\% and 1.66\%, respectively. Their combination further boosts performance by 2.19\%, highlighting these components' critical contributions to our method.

%%%%%%%%%%%%%%% method ablation description %%%%%%%%%%%%%%%%
\vspace{0.1cm}
\noindent\textbf{Verification of AFR}
The AFR module aims to balance filtering low-level domain information while preserving global structure. To evaluate this, We measure domain similarity using CKA similarity~\cite{kornblith2019similarity,Li_2021_ICCV} between the source and target domains, comparing features before and after applying AFR. As shown in Tab.~\ref{tab:feaRecon_cka}, AFR enhances domain similarity, indicating that the encoder's features effectively filter out domain-specific information while retaining more generalizable content.

\vspace{0.1cm}
\noindent\textbf{Verification of LD}
The LD module aims to reduce the encoder's reliance on the decoder for reconstruction, encouraging generalization and effective learning. As shown in Tab.~\ref{tab:feaRecon_cka}, its combination with AFR further improves domain similarity, indicating better generalization. We also ablate our design of the LD module in Tab.~\ref{tab:decoder_ablation}. It is evident that a lightweight decoder, which eliminates the MLP module and substitutes the attention-query mechanism with a cosine similarity map between tokens, achieves superior performance compared to other designs.

\begin{table}[t]
    \centering
    \resizebox{\columnwidth}{!}{
    \begin{tabular}{cccccccc}
        \toprule
        light & MLP & correlation & CropDisease & EuroSAT & ISIC2018 & ChestX & Ave. \\
        \midrule
        & \checkmark & Attn. & 94.94 & 88.56 & 47.43 & 26.77 & 64.43 \\
        \checkmark & \checkmark & Attn. & 94.94 & 88.26 & 49.47 & 27.02 & 64.93 \\
        \midrule
        \checkmark & \checkmark & Iden. & 94.87 & 88.92 & 47.60 & 26.80 & 64.55 \\
        \checkmark & \checkmark & Euc. & 94.72 & 88.18 & 50.06 & 27.09 & 65.01 \\
        \checkmark & \checkmark & Cos. & 94.70 & 88.10 & \textbf{50.35} & 27.10 & 65.06 \\ 
        \midrule
        \checkmark & & Iden. & 94.81 & 88.99 & 47.59 & 26.79 & 64.55\\
        \checkmark & & Euc. & 95.01 & \textbf{89.08} & 49.55 & 26.96 & 65.15 \\
        \checkmark & & Cos. & \textbf{95.21} & 88.90 & 49.74 & \textbf{27.10} & \textbf{65.24} \\
        \bottomrule
    \end{tabular}}
    \vspace{-0.3cm}
    \caption{Ablation study on the design of the LD module with the 5-way 5-shot accuracy.}
    % \vspace{-0.3cm}
    \label{tab:decoder_ablation}
\end{table}

\begin{figure}[t]
    \centering
    \includegraphics[width=0.95\columnwidth]{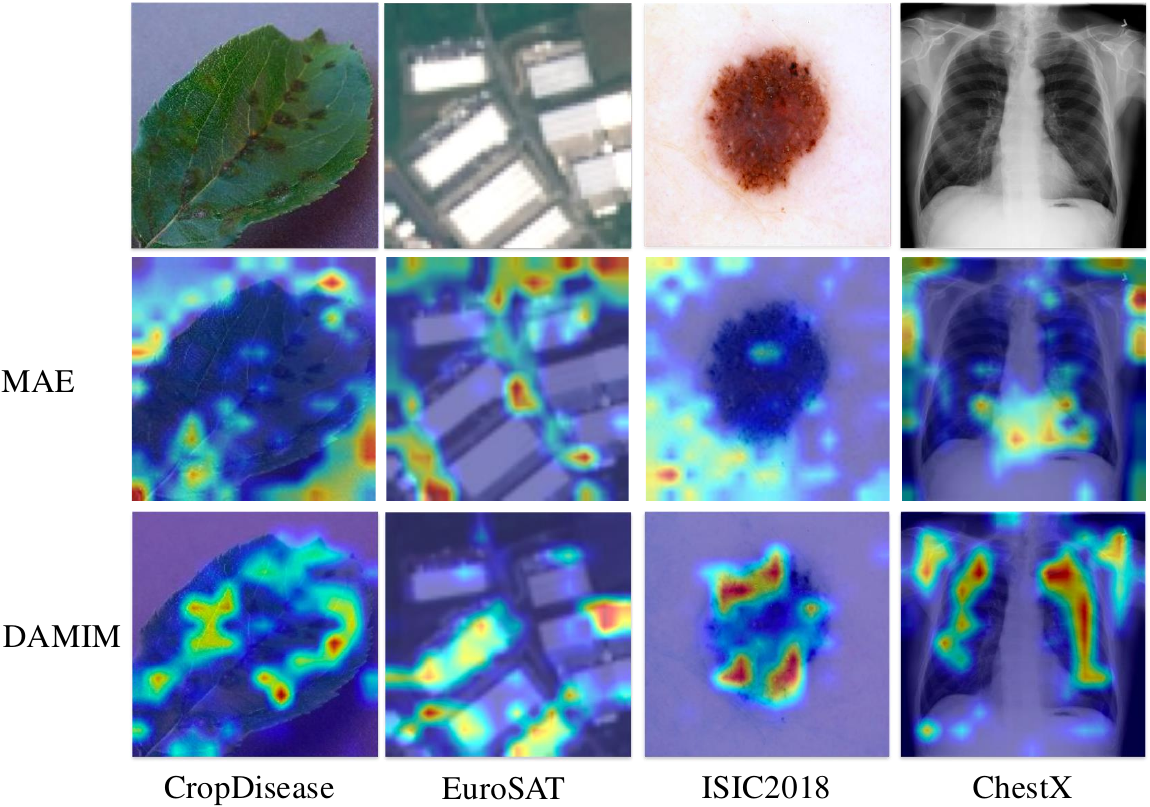}
    \caption{The heatmap for MAE and our DAMIM in four target domains. DAMIM can more effectively capture meaningful regions and comprehensive object information.}
    % \vspace{-0.4cm}
    \label{fig:visualization}
\end{figure}

\subsection{Visualization}
We generate heatmaps for MAE and our DAMIM on four target domains using Gram-CAM~\cite{Selvaraju_2017_ICCV}. As shown in Fig.~\ref{fig:visualization}, our DAMIM can more effectively capture meaningful regions and comprehensive object information, indicating that our model better generalizes the knowledge learned from the source domain to the target domain.

\section{Conclusion}
In this paper, We find that MAE tends to focus on low-level domain information during pixel reconstruction, while high-level feature reconstruction struggles with transferability, indicating a trade-off between filtering domain information and preserving global image structure. We propose DAMIM, including an Aggregated Feature Reconstruction module for balanced learning and a Lightweight Decoder module to further boost generalization. Experiments show that DAMIM outperforms state-of-the-art methods.

\section{Acknowledgments}
This work is supported by the National Natural Science Foundation of China under grants 62206102, 62436003, 62376103 and 62302184; the Science and Technology Support Program of Hubei Province under grant 2022BAA046; Hubei science and technology talent service project under grant 2024DJC078.

% \bibliography{aaai25}

\end{document}

\linenumbers*
\maketitle

\appendix

\begin{figure}
    \centering
    \includegraphics[width=0.95\columnwidth]{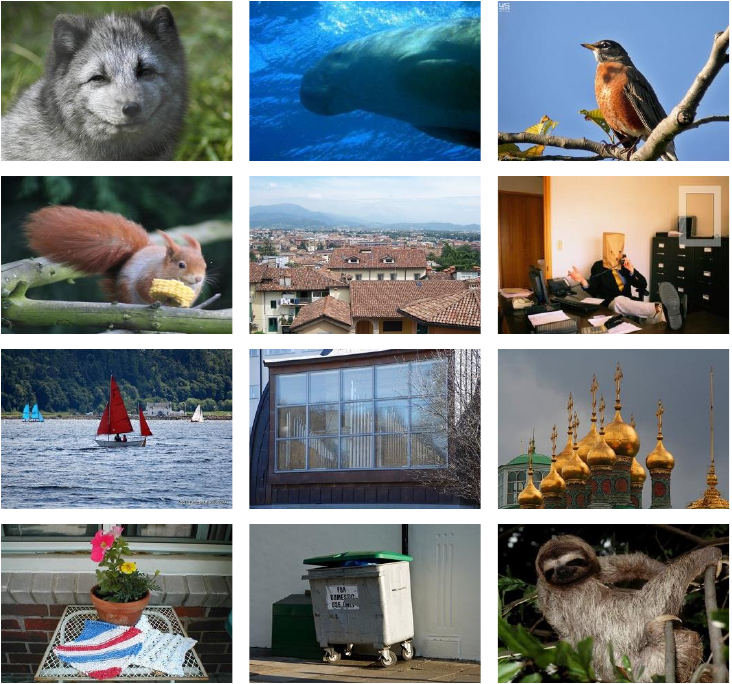}
    \caption{Samples of the miniImagenet dataset.}
    \label{fig:source_dataset}
\end{figure}

\begin{figure}
    \centering
    \includegraphics[width=0.95\columnwidth]{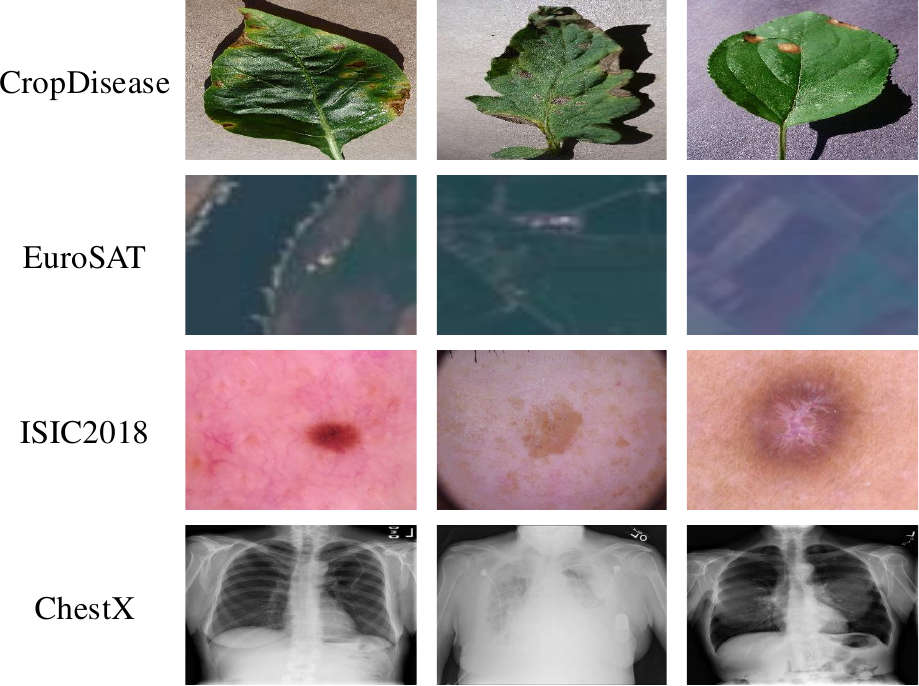}
    \caption{Samples of the CropDisease, EuroSAT, ISIC2018, and ChestX datasets.}
    \label{fig:target_dataset}
\end{figure}

\section{Dataset Description}

\noindent\textbf{\textit{mini}ImageNet}\cite{Vinyals2016Matching} is a curated subset derived from the extensive ImageNet\cite{imagenet} dataset. It includes 100 categories, each represented by 600 natural images, amounting to a total of 60,000 images. In alignment with recent studies~\cite{DBLP:journals/corr/abs-1904-04232,BSCD}, we utilize the training portion of \textit{mini}ImageNet as our source domain dataset, which comprises 64 classes and a total of 38,400 images. Representative image samples from this dataset are shown in Figure~\ref{fig:source_dataset}. Moreover, as depicted in Figure~\ref{fig:target_dataset}, we follow the approach described in~\cite{BSCD} and use datasets from four distinct domains as our target domains. These domains include plant disease images, surface satellite imagery, skin disease images, and chest X-ray images, which will be discussed in detail below.

\noindent\textbf{CropDisease}~\cite{34699834fa624a3bbc2fae48eb151339} dataset is designed for the recognition of agricultural diseases. It comprises 43,456 images spanning 19 different classes. The dataset features a variety of crops, including both healthy and diseased plants, each labeled with a specific disease category.

\noindent\textbf{EuroSAT}~\cite{helber2019eurosat} is an extensive dataset containing satellite images of Earth's surface. This dataset includes a total of 27,000 images categorized into 10 different classes, covering a wide range of geographical and topographical features.

\noindent\textbf{ISIC2018}~\cite{ISIC} is a specialized medical imaging dataset focused on the classification of skin lesions. It contains 10,015 images distributed across 7 different categories, serving as a significant resource for skin disease diagnosis.

\noindent\textbf{ChestX}~\cite{Wang_2017} is a medical imaging dataset that focuses on chest X-rays. It consists of 25,847 images categorized into 7 distinct classes, providing valuable data for chest-related disease classification tasks.

\begin{table}[t]
	\begin{center}
		\resizebox{1.0\columnwidth}{!}{
			\begin{tabular}{ccccccccc}
				\toprule
                Methods & backbone & \quad FT \quad\quad & Mark & Crop. & Euro. & ISIC. & Ches. & Ave. \\
                \midrule
                MN+AFA & ResNet10 & $\times$ & ECCV-22 & 80.07 & 69.63 & 39.88 & 23.18 & 53.19 \\
                GNN+FT & ResNet10 & $\times$ & ICLR-20 & 87.07 & 78.02 & 40.87 & 24.28 & 57.06 \\
                GNN+ATA & ResNet10 & $\times$ & IJCAI-21 & 90.59 & 83.75 & 44.91 & 24.32 & 60.39 \\
                LDP-net & ResNet10 & $\times$ & CVPR-23 & 89.40 & 82.01 & 48.06 & 26.67 & 61.29 \\
                GNN+AFA & ResNet10 & $\times$ & ECCV-22 & 88.06 & 85.58 & 46.01 & 25.02 & 61.67 \\
                SDT & ResNet10 & $\times$ & NN-24 & 90.27 & 82.02 & 48.66 & 27.20 & 62.04 \\
                FLoR & ResNet10 & $\times$ & CVPR-24 & 91.25 & 80.87 & \textbf{51.44} & 26.70 & 62.32 \\
                MEM-FS & ViT-S & $\times$ & TIP-23 & 93.74 & 86.49 & 47.38 & 26.67 & 63.57 \\
                StyleAdv & ViT-S & $\times$ & CVPR-23 & 94.85 & 88.57 & 47.73 & 26.97 & 64.53 \\
                MICM & ViT-S & $\times$ & MM-24 & 94.61 & 90.08 & 46.85 & 27.11 & 64.66 \\
                SDT & ViT-S & $\times$ & NN-24 & 95.00 & 89.60 & 47.64 & 26.72 & 64.75 \\
                FLoR & ViT-S & $\times$ & CVPR-24 & 95.28 & \textbf{90.41} & 49.52 & 26.71 & 65.48 \\
                \textbf{Ours} & ViT-S & $\times$ & \textbf{Ours} & \textbf{95.52} & 89.50 & \underline{50.76} & \textbf{27.28} & \textbf{65.77} \\
                \midrule
                Fine-tuning & ResNet10 & $\checkmark$ & ECCV-20 & 89.84 & 81.59 & 49.51 & 25.37 & 61.58 \\
                FLoR & ResNet10 & $\checkmark$ & CVPR-24 & 92.33 & 83.06 & \textbf{56.74} & 26.77 & 64.73 \\
                TACDFSL & WideResNet & $\checkmark$ & SB-22 & 93.42 & 85.19 & 45.39 & 25.32 & 62.33 \\
                PMF & ViT-S & $\checkmark$ & CVPR-22 & 92.96 & 85.98 & 50.12 & 27.27 & 64.08 \\
                StyleAdv & ViT-S & $\checkmark$ & CVPR-23 & 95.99 & 90.12 & 51.23 & 26.97 & 66.08 \\
                FLoR & ViT-S & $\checkmark$ & CVPR-24 & \textbf{96.47} & 90.75 & 53.06 & 27.02 & 66.83 \\
                \textbf{Ours} & ViT-S & $\checkmark$ & \textbf{Ours} & \underline{96.34} & \textbf{91.18} & \underline{54.86} & \textbf{27.82} & \textbf{67.55} \\
                \midrule
                ConFess\textsuperscript{*} & ResNet10 & $\checkmark$ & ICLR-2022 & 88.88 & 84.65 & 48.85 & 27.09 & 62.37 \\
                LDP-net\textsuperscript{*} & ResNet10 & $\checkmark$ & CVPR-23 & 91.89 & 84.05 & 48.44 & 26.88 & 62.82 \\
                RDC\textsuperscript{*} & ResNet10 & $\checkmark$ & CVPR-22 & 93.30 & 84.29 & 49.91 & 25.07 & 63.14 \\
                TPN+ATA\textsuperscript{*} & ResNet10 & $\checkmark$ & IJCAI-21 & 93.56 & 85.47 & 49.83 & 24.74 & 63.40 \\
                FLoR\textsuperscript{*} & ResNet10 & $\checkmark$ & CVPR-24 & 93.60 & 83.76 & \textbf{57.54} & 26.89 & 65.45 \\
                MEM-FS+RDA\textsuperscript{*} & ViT-S & $\checkmark$ & TIP-23 & 95.04 & 88.77 & 51.02 & 27.98 & 65.70 \\
                \textbf{Ours\textsuperscript{*}} & ViT-S & $\checkmark$ & \textbf{Ours} & \textbf{96.49} & \textbf{91.08} & \underline{55.44} & \textbf{28.10} & \textbf{67.78} \\
				\bottomrule
		  \end{tabular}}
    \vspace{-0.3cm}
            \caption{Comparison with more CNN-based and ViT-based methods in 5-shot. \textsuperscript{*} denotes a transductive setting.}
            \label{tab:sota_vit_5shot}
	\end{center}
\end{table}

\begin{table}[t]
	\begin{center}
		\resizebox{1.0\columnwidth}{!}{
			\begin{tabular}{ccccccccc}
            \toprule
            Method & backbone & \quad FT \quad\quad & Mark & Crop. & Euro. & ISIC. & Ches. & Ave. \\
            \midrule 
            GNN+FT & ResNet10 & $\times$ & ICLR-20 & 60.74 & 55.53 & 30.22 & 22.00 & 42.12 \\
            MN+AFA & ResNet10 & $\times$ & ECCV-22 & 60.71 & 61.28 & 32.32 & 22.11 & 44.10 \\
            GNN+ATA & ResNet10 & $\times$ & IJCAI-21 & 67.47 & 61.35 & 33.21 & 22.10 & 46.53 \\
            GNN+AFA & ResNet10 & $\times$ & ECCV-22 & 67.61 & 63.12 & 33.21 & 22.92 & 46.97 \\
            LDP-net & ResNet10 & $\times$ & CVPR-23 & 69.64 & 65.11 & 33.97 & 23.01 & 47.18 \\
            FLoR & ResNet10 & $\times$ & CVPR-24 & 73.64 & 62.90 & \textbf{38.11} & 23.11 & 49.69 \\
            SDT & ResNet10 & $\times$ & NN-24 & 73.92 & 65.87 & 36.45 & \textbf{23.22} & 49.97 \\
            MEM-FS & ViT-S & $\times$ & TIP-23 & 81.11 & 68.11 & 32.97 & 22.76 & 51.24 \\
            StyleAdv & ViT-S & $\times$ & CVPR-23 & 81.22 & 72.15 & 33.05 & 22.92 & 52.34 \\
            SDT & ViT-S & $\times$ & NN-24 & 81.03 & 72.71 & 33.40 & 22.79 & 52.48 \\
            FLoR & ViT-S & $\times$ & CVPR-24 & 81.81 & 72.39 & 34.20 & 22.78 & 52.80 \\
            \textbf{Ours} & ViT-S & $\times$ & \textbf{Ours} & \textbf{82.34} & \textbf{72.87} & 34.66 & 22.97 & \textbf{53.21} \\
            \midrule
            Fine-tuning & ResNet10 & $\checkmark$ & ECCV-20 & 73.43 & 66.17 & 34.60 & 22.13 & 49.08 \\
            FLoR & ResNet10 & $\checkmark$ & CVPR-24 & 84.04 & 69.13 & \textbf{38.81} & 23.12 & 53.78 \\
            PMF & ViT-S & $\checkmark$ & CVPR-22 & 80.79 & 70.74 & 30.36 & 21.73 & 50.91 \\
            FLoR & ViT-S & $\checkmark$ & CVPR-24 & 83.55 & 73.09 & 35.49 & 23.26 & 53.85 \\
            StyleAdv & ViT-S & $\checkmark$ & CVPR-23 & \textbf{84.11} & \textbf{74.93} & 33.99 & 22.92 & 53.99 \\
            \textbf{Ours} & ViT-S & $\checkmark$ & \textbf{Ours} & \underline{83.90} & \underline{73.61} & \underline{36.35} & \textbf{23.38} & \textbf{54.31} \\
            \midrule
            LDP-net\textsuperscript{*} & ResNet10 & $\checkmark$ & CVPR-23 & 81.24 & 73.25 & 33.44 & 22.21 & 52.54 \\
            TPN+ATA\textsuperscript{*} & ResNet10 & $\checkmark$ & IJCAI-21 & 82.47 & 70.84 & 35.55 & 22.45 & 52.83 \\
            RDC\textsuperscript{*} & ResNet10 & $\checkmark$ & CVPR-22 & 85.79 & 70.51 & 36.28 & 22.32 & 53.73 \\
            FLoR\textsuperscript{*} & ResNet10 & $\checkmark$ & CVPR-24 & 86.30 & 71.38 & \textbf{41.67} & 23.12 & 55.62 \\
            MEM-FS+RDA\textsuperscript{*} & ViT-S & $\checkmark$ & TIP-23 & 83.74 & 75.91 & 37.07 & 23.85 & 55.14 \\
            \textbf{Ours\textsuperscript{*}} & ViT-S & $\checkmark$ & \textbf{Ours} & \textbf{86.74} & \textbf{77.23} & \underline{38.07} & \textbf{23.91} & \textbf{56.49} \\
            \bottomrule
		  \end{tabular}}
    \vspace{-0.3cm}
            \caption{Comparison with more CNN-based and ViT-based methods in 1-shot. \textsuperscript{*} denotes a transductive setting.}
            \label{tab:sota_vit_1shot}
    \end{center}
\end{table}

\begin{table}[t]
    \resizebox{1.0\columnwidth}{!}{
        \centering
        \begin{tabular}{ccccccc}
        \toprule
        Method & Shot & ChestX & ISIC2018 & EuroSAT & CropDisease & Average \\
        \midrule
        iBOT & 1 & 22.69 & 31.54 & 70.74 & 81.58 & 51.64 \\
        \textbf{iBOT+Ours} & 1 & \textbf{23.19} & \textbf{34.92} & \textbf{73.52} & \textbf{83.68} & \textbf{53.83}\\
        \midrule
        iBOT & 5 & 26.44 & 44.28 & 87.17 & 94.74 & 63.16\\
        \textbf{iBOT+Ours} & 5 & \textbf{27.12} & \textbf{49.20} & \textbf{89.48} & \textbf{95.48} & \textbf{65.32} \\
        \bottomrule
        \end{tabular}}
    \vspace{-0.3cm}
    \caption{Our method with iBOT-pretrained ViT-S}
    \label{tab:ibot_pretrain}
\end{table}

\section{Comparison with more CNN methods}
As depicted in Tab.~\ref{tab:sota_vit_5shot} and Tab.~\ref{tab:sota_vit_1shot}, we comprehensively compare various ViT and CNN-based approaches on CDFSL tasks. Our methods consistently outperform all others, achieving the best performance. These results underscore the effectiveness of our approach.

\section{Applying Our Method to ViT Variants}

\begin{table}[t]
    \resizebox{1.0\columnwidth}{!}{
        \centering
        \begin{tabular}{ccccccc}
        \toprule
        Method & Shot & ChestX & ISIC2018 & EuroSAT & CropDisease & Average \\
        \midrule
        DINO-B & 1 & 22.67 & 34.08 & 69.69 & 82.57 & 52.25 \\
        \textbf{DINO-B+Ours} & 1 & \textbf{22.93} & \textbf{35.58} & \textbf{72.97} & \textbf{84.83} & \textbf{54.08}\\
        \midrule
        DINO-B & 5 & 26.26 & 48.77 & 87.71 & 95.25 & 64.50\\
        \textbf{DINO-B+Ours} & 5 & \textbf{26.56} & \textbf{50.09} & \textbf{90.07} & \textbf{95.85} & \textbf{65.64} \\
        \bottomrule
        \end{tabular}}
        \vspace{-0.3cm}
    \caption{Our method with DINO-pretrained ViT-B}
    \label{tab:dino_base}
\end{table}

\begin{table}[t]
    \resizebox{1.0\columnwidth}{!}{
        \centering
        \begin{tabular}{cccccc}
        \toprule
        Method & ChestX & ISIC2018 & EuroSAT & CropDisease & Average \\
        \midrule
        IWG & 26.54 & 49.49 & 89.55 & 94.08 & 65.09 \\
        IOG & 26.56 & 49.38 & 89.52 & 95.06 & 65.13 \\
        IE & 26.58 & 49.32 & 89.48 & 95.08 & 65.12 \\
        SOG & 27.09 & 49.92 & 89.35 & 95.29 & 65.41 \\
        \textbf{Ours} & \textbf{27.28} & \textbf{50.76} & 89.50 & \textbf{95.52} & \textbf{65.77}\\
        \bottomrule
        \end{tabular}}
        \vspace{-0.3cm}
    \caption{Auxilary Encoder design}
    \label{tab:encoder_design}
\end{table}

We also apply our method to the ViT-Small model pretrained using iBOT\cite{zhou2022image} and the ViT-Base model pretrained using DINO\cite{caron2021emerging}. iBOT is a self-supervised pre-training framework that learns semantic representations of images through Masked Image Modeling (MIM) and an online tokenizer, enabling effective pre-training of vision Transformers without requiring labeled data. The results are shown in Table~\ref{tab:ibot_pretrain}. Here, iBOT represents the baseline ViT model pretrained with iBOT, while iBOT+Ours indicates our method applied to the iBOT-pretrained ViT. Our method achieves a notable improvement on the ViT pretrained with iBOT, with a 2.19-point increase in 1-shot and a 2.16-point increase in 5-shot. The results of applying our method to the DINO-pretrained ViT-Base are presented in Table~\ref{tab:dino_base}. DINO-B represents the baseline ViT-Base model pretrained with DINO in our CDFSL task, and DINO-B+Ours indicates our method applied to the DINO-pretrained ViT-Base. This also shows an improvement, with a 1.83-point increase in 1-shot and a 1.14-point increase in 5-shot.

\section{Auxilary Encoder}
We also design various auxiliary encoder architectures: Two Independent Encoders with Grad (IWG), Two Independent Encoders without Grad (IOG), Two Independent Encoders with EMA (IE), Shared Encoders without Grad (SOG), and our Shared Encoders with Grad while detaching layer features. As depicted in Table~\ref{tab:encoder_design}, our shared encoder performs best.

\section{Centered Kernel Alignment}
Centered Kernel Alignment (CKA)~\cite{kornblith2019similarity} is a statistical method for measuring the similarity between representations learned by different neural networks or between different layers within the same network. Originally developed for kernel methods, CKA is effective in comparing high-dimensional representations.
To compute CKA between two sets of data representations $X \in \mathbb{R}{n \times p}$ and $Y \in \mathbb{R}{n \times p}$, we first calculate the Gram matrices $K = XX^\top$ and $L = YY^\top$. These matrices capture the inner products between all pairs of data points in their respective feature spaces. Next, we center the Gram matrices using:
\begin{equation}
    \begin{aligned}
    & K_c = HKH \\
    & L_c = HLH
    \end{aligned}
\end{equation}
where $H = I_n - \frac{1}{n}\mathbf{1}_n \mathbf{1}_n^\top$ is the centering matrix, $I_n$ is the identity matrix, and $\mathbf{1}_n$ is a vector of ones.Finally, the centered kernel alignment is computed as:
\begin{equation}
CKA(K,L) = \frac{\text{Tr}(K_cL_c)}{\sqrt{\text{Tr}(K_c^2)\text{Tr}(L_c^2)}}
\end{equation}
where \text{Tr} denotes the trace of a matrix.

CKA is particularly useful for measuring the similarity~\cite{kornblith2019similarity,Li_2021_ICCV} between different datasets or network layers. High CKA values indicate strong similarity, while low values suggest significant differences. In this paper, we use CKA to evaluate the similarity of features across different network layers. Specifically, for each dataset, we extract features from the model and compute the CKA similarity between the source and target domains to understand how well the model generalizes across different data distributions.

% \bibliography{aaai25}